%% file: main.tex
\documentclass{article}

\usepackage{microtype}
\usepackage{graphicx}
\usepackage{subcaption}
\usepackage[table]{xcolor}
\usepackage{booktabs} 
\usepackage{fontawesome5}

\usepackage{hyperref}

\usepackage[preprint]{icml2026}

\usepackage{amsmath}
\usepackage{amssymb}
\usepackage{mathtools}
\usepackage{amsthm}

\usepackage[capitalize,noabbrev]{cleveref}

\theoremstyle{plain}

\theoremstyle{definition}

\theoremstyle{remark}

\usepackage[textsize=tiny]{todonotes}

\usepackage{enumitem}
\setlist[itemize]{noitemsep, topsep=0pt}
\setlist[enumerate]{noitemsep, topsep=0pt}

\newcommand{\website}{https://chenshuo20.github.io/Context\_Forcing}
\newcommand{\code}{https://github.com/TIGER-AI-Lab/Context-Forcing}
\newcommand{\linkcolor}{\textcolor[RGB]{230,0,115}}

\icmltitlerunning{Context Forcing: Consistent Autoregressive Video Generation with Long Context}

\begin{document}

\newsavebox{\teaserimage}
\sbox{\teaserimage}{%
  \includegraphics[width=\textwidth]{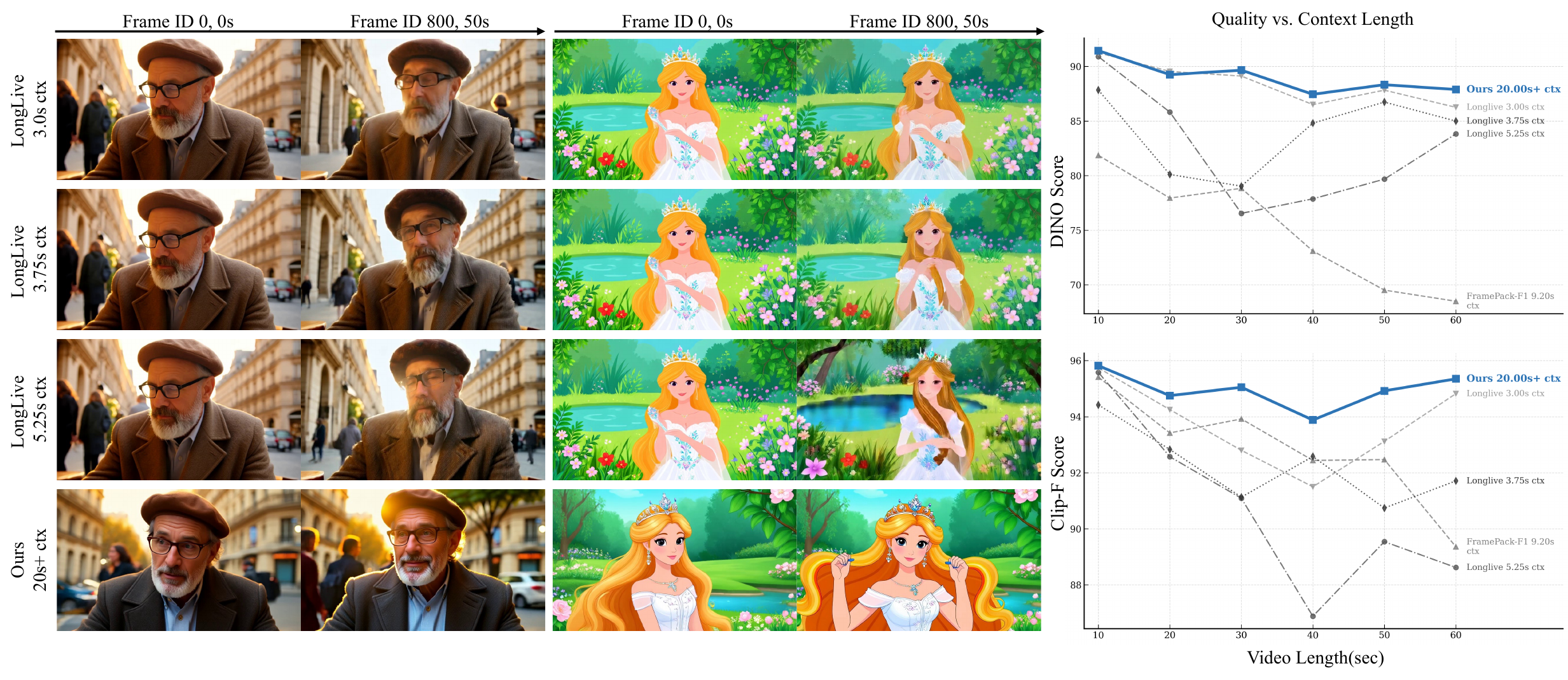}
}

\twocolumn[
  \icmltitle{Context Forcing: Consistent Autoregressive Video Generation with \\ Long Context}



    \icmlsetsymbol{equal}{*}
    
    \begin{icmlauthorlist}
      \icmlauthor{Shuo Chen}{equal,ucm}
      \icmlauthor{Cong Wei}{equal,waterloo}
      \icmlauthor{Sun Sun}{waterloo}
      \icmlauthor{Ping Nie}{waterloo}
      \icmlauthor{Kai Zhou}{netmind}
      \icmlauthor{Ge Zhang}{map}
      \icmlauthor{Ming-Hsuan Yang}{ucm}
      \icmlauthor{Wenhu Chen}{waterloo}
    \end{icmlauthorlist}
    
    \icmlaffiliation{ucm}{Department of EECS, University of California, Merced, USA}
    \icmlaffiliation{waterloo}{University of Waterloo, Canada}
    \icmlaffiliation{netmind}{Netmind.AI}
    \icmlaffiliation{map}{M-A-P}

  \icmlcorrespondingauthor{Ming-Hsuan Yang}{mhyang@ucmerced.edu}
  \icmlcorrespondingauthor{Wenhu Chen}{wenhuchen@uwaterloo.ca}

  \icmlkeywords{Machine Learning, ICML}

  \vskip 0.3in

\begin{center}
    \faGlobe\ \textbf{Website: }
    \href{\website}{\linkcolor{\texttt{\website}}}
  \end{center}

  \begin{center}
    \faGithub\ \textbf{Code:}
    \href{\code}{\linkcolor{\texttt{\code}}}
    \vspace{1.0em} 
  \end{center}

\begin{center}
    \usebox{\teaserimage}
    \vspace{-3mm}
    \captionof{figure}{
    \textbf{Context Forcing mitigates the forgetting--drifting dilemma.}
    (1) State-of-the-art models are limited by short context windows (3.0--9.2\,s), which leads to poor long-term consistency (\emph{Forgetting}).
    (2) For streaming long-context tuning baselines (e.g., LongLive), enlarging the context window during inference (3.0 $\rightarrow$ 5.25\,s) causes error accumulation and distribution shift (\emph{Drifting}).
    In contrast, \textbf{Context Forcing} supports \textbf{20s+} context while maintaining strong long-term consistency.
    }
    \label{fig:dilemma}
  \end{center}
  \vspace{1.5em}
]



\printAffiliationsAndNotice{\icmlEqualContribution}

\input{sec/0_abstract}

\input{sec/1_introduction}

\input{sec/2_related_works}

\input{sec/3_methodology}

\input{sec/4_experiments}
\input{sec/5_conclusion}
\input{sec/6_impact_statement}
\bibliography{main}
\bibliographystyle{icml2026}
\input{sec/X_suppl}

\end{document}

%% file: sec/0_abstract.tex
\begin{abstract}
Recent approaches to real-time long video generation typically employ streaming tuning strategies, attempting to train a long-context student using a short-context (memoryless) teacher. In these frameworks, the student performs long rollouts but receives supervision from a teacher limited to short 5-second windows. This structural discrepancy creates a critical \textbf{student-teacher mismatch}: the teacher's inability to access long-term history prevents it from guiding the student on global temporal dependencies, effectively capping the student's context length.
To resolve this, we propose \textbf{Context Forcing}, a novel framework that trains a long-context student via a long-context teacher. By ensuring the teacher is aware of the full generation history, we eliminate the supervision mismatch, enabling the robust training of models capable of long-term consistency. To make this computationally feasible for extreme durations (e.g., 2 minutes), we introduce a context management system that transforms the linearly growing context into a \textbf{Slow-Fast Memory} architecture, significantly reducing visual redundancy.
Extensive results demonstrate that our method enables effective context lengths exceeding 20 seconds—$2\text{--}10\times$ longer than state-of-the-art methods like LongLive and Infinite-RoPE. By leveraging this extended context, Context Forcing preserves superior consistency across long durations, surpassing state-of-the-art baselines on various long video evaluation metrics.
\end{abstract}

%% file: sec/1_introduction.tex
\section{Introduction}

In recent years, video diffusion models based on architectures such as the Denoising Diffusion Transformer(DiT)~\cite{peebles2023scalable} have achieved remarkable success in generating photorealistic videos~\cite{wan2025wan}. While bidirectional models perform well for short clips, their computational cost limits long-form generation. To address this, the field is moving toward causal video architectures~\cite{yin2024slow,huang2025self}, which, like Large Language Models, can theoretically generate infinite-length videos by predicting future frames from past context.

Despite this promise, current causal video models struggle to maintain coherence over long-term contexts. Effective context is often limited to just a few seconds~\cite{cui2025self, yang2025longlive, zhang2025packing, huang2025self, yesiltepe2025infinity}, beyond which identity shifts and temporal inconsistencies emerge. We identify the root cause as a fundamental \textbf{student-teacher mismatch}. As illustrated in Figure~\ref{fig:compare_vis}(b), current methods typically train a student to perform long rollouts using supervision from a memoryless teacher limited to short windows (e.g., 5 seconds). The teacher’s inability to access long-term history prevents it from guiding the student on global temporal dependencies, effectively capping the student’s learnable context length.

This mismatch results in a critical challenge for real-time long-context video generation, which we term the \emph{Forgetting-Drifting Dilemma} (Figure~\ref{fig:dilemma}). Existing methods face an unavoidable trade-off:
\begin{itemize}
    \item \textbf{Forgetting:} Restricting the model to a short memory window minimizes error accumulation but causes the model to lose track of previous subjects and scenes during long rollout.
    \item \textbf{Drifting:} Maintaining a long context preserves identity but exposes the model to its own accumulated errors. Without a teacher capable of correcting these long-term deviations, the video distribution progressively drifts away from the real manifold.
\end{itemize}

To address these challenges, we propose \textbf{Context Forcing}, a framework that distills a long-context teacher into a long-context student. Our approach resolves the context-drifting dilemma by bridging the capability gap between teacher and student. We first leverage a Context Teacher pretrained on video continuation tasks, which is capable of processing long-context inputs. This teacher guides the student via \emph{Contextual Distribution Matching Distillation}, explicitly transferring the ability to model long-term dependencies and ensuring global consistency. Furthermore, by exposing the student to imperfect, self-generated contexts during training, we enable it to actively recover from accumulated artifacts. The resulting robustness allows for $2-10\times$ longer duration Key-Value (KV) cache management (maintaining 20+ seconds of history) compared to prior SOTA (1.5--9.2 seconds of history) during inference, effectively addressing the forgetting-drifting trade-off and enabling consistent, long-form video generation.

The contributions of this work are:
\begin{itemize}
    \item We introduce \textbf{Context Forcing}, a novel framework that mitigates the student-teacher mismatch in training real-time long video models. By distilling from a long-context teacher aware of the full generation history, we enable the robust training of a long-context student capable of long-term consistency.

    \item To support this, we design a context management system that transforms the linearly growing context into a \textbf{Slow-Fast Memory} architecture, significantly reducing visual redundancy. This mechanism enables effective context lengths exceeding 20 seconds—$2\text{--}10\times$ longer than state-of-the-art methods.

    \item We demonstrate that, equipped with these extended context lengths, our model preserves superior consistency across long durations, surpassing state-of-the-art baselines on various long video evaluation metrics.
\end{itemize}

%% file: sec/2_related_works.tex
\section{Related Work}

\begin{figure*}[tp]
    \centering
    \includegraphics[width=\linewidth]{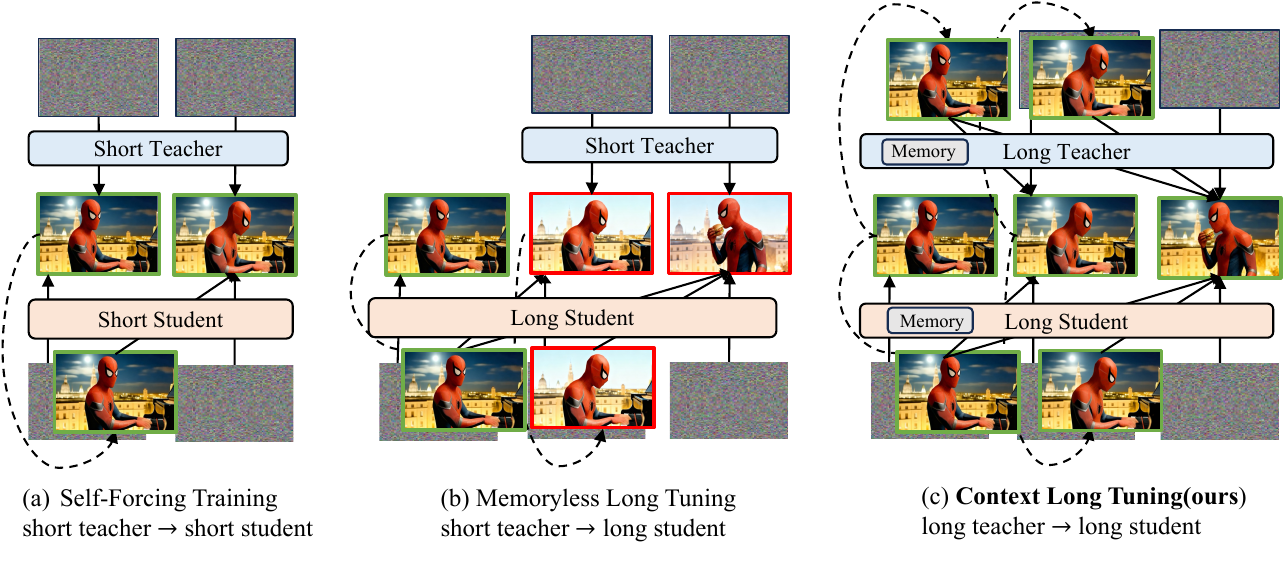}
    \vspace{-5mm}
    \caption{\textbf{Training paradigms for AR video diffusion models.} (a) Self-forcing: A student matches a teacher capable of generating only 5s video using a 5s self-rollout. (b) Longlive~\citep{yang2025longlive}: The student performs long rollouts supervised by a memoryless 5s teacher on random chunks. The teacher's inability to see beyond its 5s window creates a student-teacher mismatch. (c) \textbf{Context Forcing (Ours)}: The student is supervised by a long-context teacher aware of the full generation history, resolving the mismatch in (b).}
    \label{fig:compare_vis}
    \vspace{-3mm}
\end{figure*}

 \noindent \textbf{Long Video Generation.}  
The high computational cost of Diffusion Transformers (DiTs)~\cite{kong2024hunyuanvideo, wan2025wan, peebles2023scalable, yang2024cogvideox} has limited video generation to short clips. To extend temporal horizons, many works combine diffusion with autoregressive (AR) prediction~\cite{kim2024fifo, lin2025autoregressive, gu2025long}, including NOVA~\cite{deng2024autoregressive}, Pyramid-Flow~\cite{jin2024pyramidal}, and MAGI-1~\cite{teng2025magi}. Other approaches improve efficiency via causal or windowed attention and KV caching~\cite{yin2024slow, huang2025self, kodaira2025streamdit}, or extend context through training-free positional encoding modifications~\cite{lu2024freelong, lu2025freelong++, zhao2025riflex}. However, most methods still struggle with global consistency beyond 10-20 seconds.   A key challenge of long video generation is error accumulation (drifting), addressed either during training by exposing models to drifted inputs~\cite{cui2025self, chen2024diffusion, chen2025skyreels} or during inference via recaching, sampling strategies, or feedback~\cite{yang2025longlive, zhang2025packing, li2025stable}. To enable real-time generation, recent works distill multi-step diffusion into few-step models~\cite{valevski2024diffusion, liu2023instaflow, luo2023lcm, sauer2024fast}, including Distribution Matching Distillation (DMD/DMD2)~\cite{yin2024one, yin2024improved,wang2023prolificdreamer} and Consistency Models (CM)~\cite{song2023consistency, wang2024phased}.

\noindent\textbf{Causal Video Generation.} Causal video generation synthesizes video sequences under strict temporal ordering constraints, thereby enabling streaming inference and long-horizon synthesis. 
Although early autoregressive models~\cite{vondrick2016generating,kalchbrenner2017video} generated frames or tokens sequentially, they often suffered from error accumulation and poor scalability. Recent diffusion-based frameworks have improved visual fidelity by incorporating causal architectural priors, such as the block-wise causal attention introduced in CausVid~\cite{yin2024slow}. To mitigate distribution shift, Self-Forcing~\cite{huang2025self}, LongLive~\cite{yang2025longlive} and Self-Forcing++~\cite{cui2025self} align training with inference by conditioning on prior outputs via KV caching and rollout-based objectives. InfinityRoPE~\cite{yesiltepe2025infinity} achieve a reduction of error accumulation by modifying positional encodings. Further research has addressed efficient long-context inference through windowed attention, as seen in StreamDiT~\cite{kodaira2025streamdit}.

\noindent \textbf{Memory Mechanism for Video Generation}  
Memory mechanisms are key to extending temporal context and maintaining consistency in long-horizon generation. WorldPlay~\cite{sun2025worldplay}, Context as Memory~\cite{yu2025context}, and WorldMem~\cite{xiao2025worldmem} and Framepack~\cite{zhang2025packing} introduce explicit memory structures to accumulate scene or contextual information over time, while RELIC~\cite{hong2025relic} employs recurrent latent states for efficient long-range dependency modeling. PFP~\cite{zhang2026pretrainingframepreservationautoregressive} compress long videos into short context by training a novel compression module.

%% file: sec/3_methodology.tex


\section{Methodology}

We operate within the causal autoregressive framework, where the generation of a long video $X_{1:N}$ is decomposed into a sequence of conditional steps over frames or short chunks $X_t$. State-of-the-art methods, such as CausVid~\cite{yin2024slow} and Self-Forcing~\cite{huang2025self}, enforce strict temporal causality via block-wise attention, modeling the distribution as $\prod_t p(X_t \mid X_{<t})$. These approaches typically employ Distribution Matching Distillation (DMD)~\cite{yin2024one} to distill a high-quality bidirectional teacher into a causal student. 
Building on these foundations, we introduce \textbf{Context Forcing}.

Our goal is to train a causal video diffusion model, parameterized by $\theta$, whose induced distribution over \emph{long videos} $p_\theta(X_{1:N})$ matches the real data distribution $p_{\text{data}}(X_{1:N})$. Here, $N$ represents a duration spanning tens or hundreds of seconds. The objective is to minimize the global long-horizon KL divergence:

\begin{figure*}[tp]
    \centering
    \includegraphics[width=\linewidth]{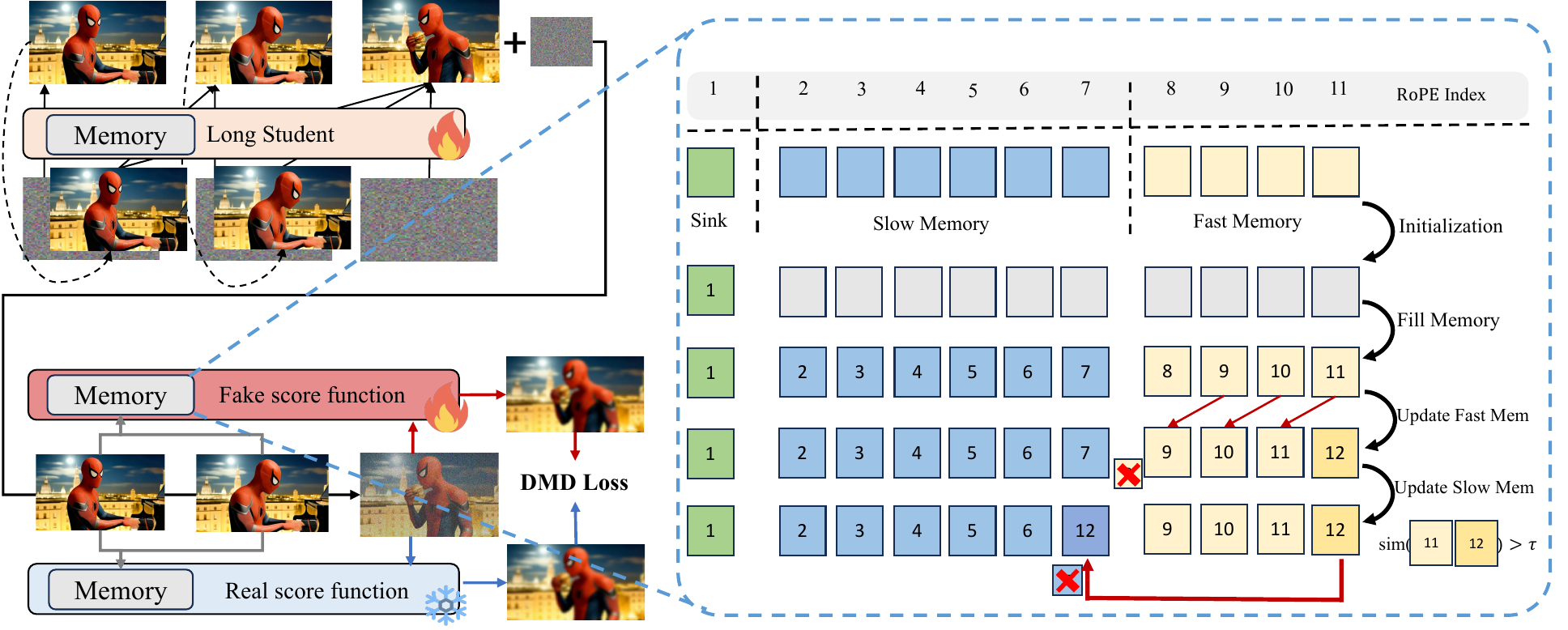}
        \vspace{-2mm}
    \caption{\textbf{Context Forcing and Context Management System.} We use KV Cache as the context memory, and we organize it into three parts: sink, slow memory and fast memory. During contextual DMD training, the long teacher provides supervision to the long student by utilizing the same context memory mechanism.}
    \label{fig:pipeline}
    \vspace{-3mm}
\end{figure*}

\begin{equation}
\label{eq:global-kl}
\mathcal{L}_{\text{global}} = \min_{\theta} \mathrm{KL}\big(p_\theta(X_{1:N})\;\|\;p_{\text{data}}(X_{1:N})\big).
\end{equation}
Directly optimizing Eq.~\eqref{eq:global-kl} ensures long-term coherence but is computationally intractable for large $N$. By applying the chain rule of KL divergence, we decompose the global objective into two components:
\begin{equation}
\label{eq:kl-decomposition}
\begin{aligned}
& \mathcal{L}_{\text{global}} ={} \underbrace{\mathrm{KL}\big(p_\theta(X_{1:k}) \,\|\, p_{\text{data}}(X_{1:k})\big)}_{\mathcal{L}_{\text{local}}\text{: Local Dynamics}} \\
& + \underbrace{\mathbb{E}_{X_{1:k} \sim p_\theta} \Big[ \mathrm{KL}\big(p_\theta(X_{k+1:N} | X_{1:k}) \,\|\, p_{\text{data}}(X_{k+1:N} | X_{1:k})\big) \Big]}_{\mathcal{L}_{\text{context}}\text{: Global Continuation Dynamics}}
\end{aligned}
\end{equation}
This decomposition motivates our two-stage curriculum:
\begin{itemize}
    \item \textbf{Stage 1 (Optimizing $\mathcal{L}_{\text{local}}$):} We match the distribution of short windows ($X_{1:k}$) to the real data distribution to learn local dynamics.
    \item \textbf{Stage 2 (Optimizing $\mathcal{L}_{\text{context}}$):} We match the model's continuation predictions ($X_{k+1:N}$) with the temporal evolution of real data to learn long-term dependencies.
\end{itemize}

\subsection{Stage 1: Local Distribution Matching}
\label{sec:stage1}

The first stage warms up the causal student by minimizing $\mathcal{L}_{\text{local}}$. Given a teacher distribution $p_T(X_{1:k})$ (approximately the real data), we optimize:
\begin{equation}
\mathcal{L}_{\text{local}} = \mathrm{KL}\big(p_\theta(X_{1:k})\,\|\,p_T(X_{1:k})\big),
\end{equation}
where $k$ corresponds to a 1--5 second window. We estimate the distribution matching gradient follow DMD~\citep{yin2024one}. Let $x = G_\theta(z)$ for noise $z$, and let $x_t$ be the diffused version of $x$ at timestep $t$. The gradient is given by:
\begin{equation}
\label{eq:stage1_grad}
\nabla_\theta \mathcal{L}_{\text{local}}
\approx
\mathbb{E}_{z, t, x_t}
\Big[
w_t \alpha_t \,
\big(s_\theta(x_t, t) - s_T(x_t, t)\big)\,
\frac{\partial G_\theta(z)}{\partial \theta}
\Big],
\end{equation}
where $s_\theta$ and $s_T$ are the student and teacher scores, respectively, and $w_t$ is a weighting function. This stage ensures $p_\theta(X_{1:k}) \approx p_{\text{data}}(X_{1:k})$, providing high-quality contexts for the subsequent stage.

\vspace{-2mm}

\subsection{Stage 2: Contextual Distribution Matching}
\label{sec:cdmd}

Stage 2 targets $\mathcal{L}_{\text{context}}$, the second term of Eq.~\eqref{eq:kl-decomposition}. This term requires minimizing the divergence between the student's continuation $p_\theta(\cdot | X_{1:k})$ and the true data continuation $p_{\text{data}}(\cdot | X_{1:k})$.

However, $p_{\text{data}}$ is not directly accessible for arbitrary contexts generated by the student. To solve this, we employ a pretrained \textbf{Context Teacher} $T$, which provides a reliable proxy distribution $p_T(X_{k+1:N} \mid X_{1:k})$. We rely on two key assumptions to justify using the teacher as a target:

\noindent\textbf{Assumption 1 (Teacher reliability near student contexts).}
\textit{Whenever the student context $X_{1:k} \sim p_\theta(X_{1:k})$ remains close to the real data manifold, the teacher's continuation $p_T(X_{k+1:N} \mid X_{1:k})$ is accurate.} This holds whenever the teacher is well-trained on real video prefixes.

\noindent\textbf{Assumption 2 (Approximate real prefixes).}
\textit{Stage 1 successfully aligns $p_\theta(X_{1:k})$ with $p_{\text{data}}(X_{1:k})$.} This ensures that student rollouts remain within the teacher's reliable region during Stage 2 training.

\begin{figure*}[tp]
    \centering
    \includegraphics[width=\linewidth]{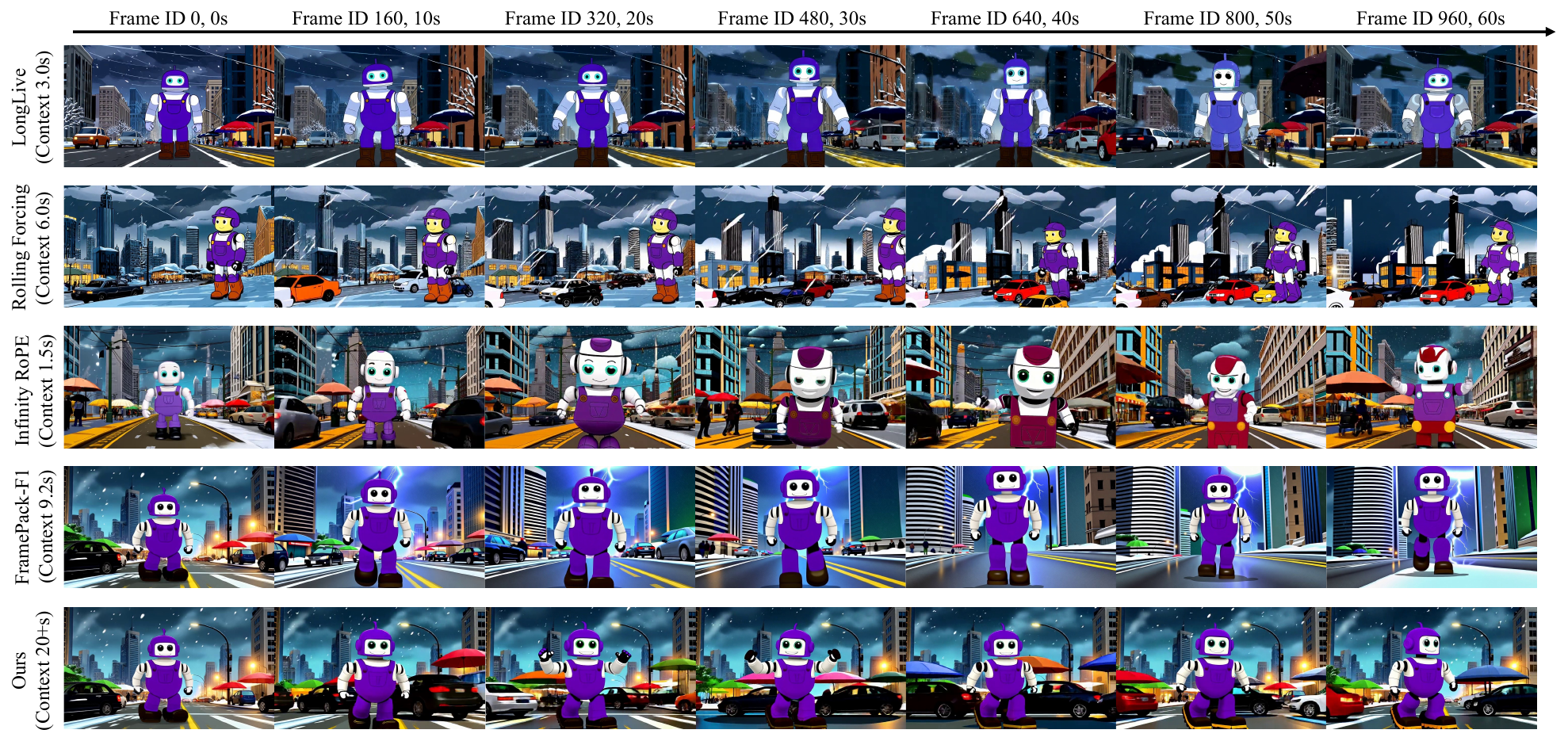}
    \caption{\textbf{Comparison on 1-min Video Generation.} Our method keeps both the background and subject consistent across 1-min video, while other baselines have different levels drifting or identity shift.}
\end{figure*}

Under these assumptions, we approximate $p_{\text{data}} \approx p_T$ in the second term of Eq.~\eqref{eq:kl-decomposition}, yielding the \textbf{Contextual DMD (CDMD)} objective:
\begin{equation}
\label{eq:cdmd_obj}
\begin{aligned}
\mathcal{L}_{\text{CDMD}} ={} & \mathbb{E}_{X_{1:k} \sim p_\theta(X_{1:k})} \\
& \Big[ \mathrm{KL}\big(p_\theta(X_{k+1:N} \mid X_{1:k})\,\|\,p_T(X_{k+1:N} \mid X_{1:k})\big) \Big]
\end{aligned}
\end{equation}
Crucially, the expectation is over $X_{1:k} \sim p_\theta$, ensuring the student is trained on its \emph{own} rollouts, thereby mitigating exposure bias.

\noindent \textbf{Score-based CDMD Gradient.}
We estimate the gradient of Eq.~\eqref{eq:cdmd_obj} using a conditional variant of the DMD gradient. Let $x_{\text{cont}} = G_\theta(z_{\text{cont}} \mid X_{1:k})$ be the generated continuation, and $x_{t, \text{cont}}$ be its diffused version. Running both fake score and real score models on the \emph{same} student-generated context produces scores $s_\theta(\cdot \mid X_{1:k})$ and $s_T(\cdot \mid X_{1:k})$. The gradient is:
\begin{equation}
\label{eq:cdmd_grad}
\begin{aligned}
&  \nabla_\theta \mathcal{L}_{\text{CDMD}} \approx{} \mathbb{E}_{\substack{X_{1:k} \sim p_\theta \\ z_{\text{cont}}, t}} \Big[ w_t \alpha_t \big( s_\theta(x_{t, \text{cont}}, t \mid X_{1:k}) \\ 
& - s_T(x_{t, \text{cont}}, t \mid X_{1:k}) \big) \frac{\partial G_\theta(z_{\text{cont}} \mid X_{1:k})}{\partial \theta} \Big].
\end{aligned}
\end{equation}
By descending Eq.~\eqref{eq:cdmd_grad}, we align the student's long-term autoregressive dynamics with the teacher's robust priors.

\paragraph{Long Self-Rollout Curriculum.}
Minimizing $\mathcal{L}_{\text{context}}$ requires the context horizon $k$ to approach the full sequence length $N$. However, sampling $X_{1:k} \sim p_\theta$ for large $k$ early in training causes severe distribution shift due to accumulated drift. To mitigate this, we employ a dynamic horizon schedule $N_{\max}^{(t)}$ that grows linearly with training step $t$. At each iteration, the rollout length is sampled as $k \sim \mathcal{U}(k_{\min}, N_{\max}^{(t)})$. This curriculum initializes training in the stable Stage~1 regime ($k \approx k_{\min}$) and progressively exposes the model to long-range dependencies.

\paragraph{Clean Context Policy.}
Self Forcing~\cite{huang2025self} typically generates rollouts using a random timestep selection strategy to ensure supervision across all diffusion steps. We retain this random exit policy for the \emph{target} frames $X_{k+1:N}$ to preserve gradient coverage, but enforce that the \emph{context} frames $X_{1:k}$ are fully denoised. We apply a complete few-step denoising process to the context. This decoupling ensures the context remains informative and aligned with the teacher's training distribution but also maintains supervision for every diffusion step.

\begin{figure*}[tp]
    \centering
    \includegraphics[width=0.95\linewidth]{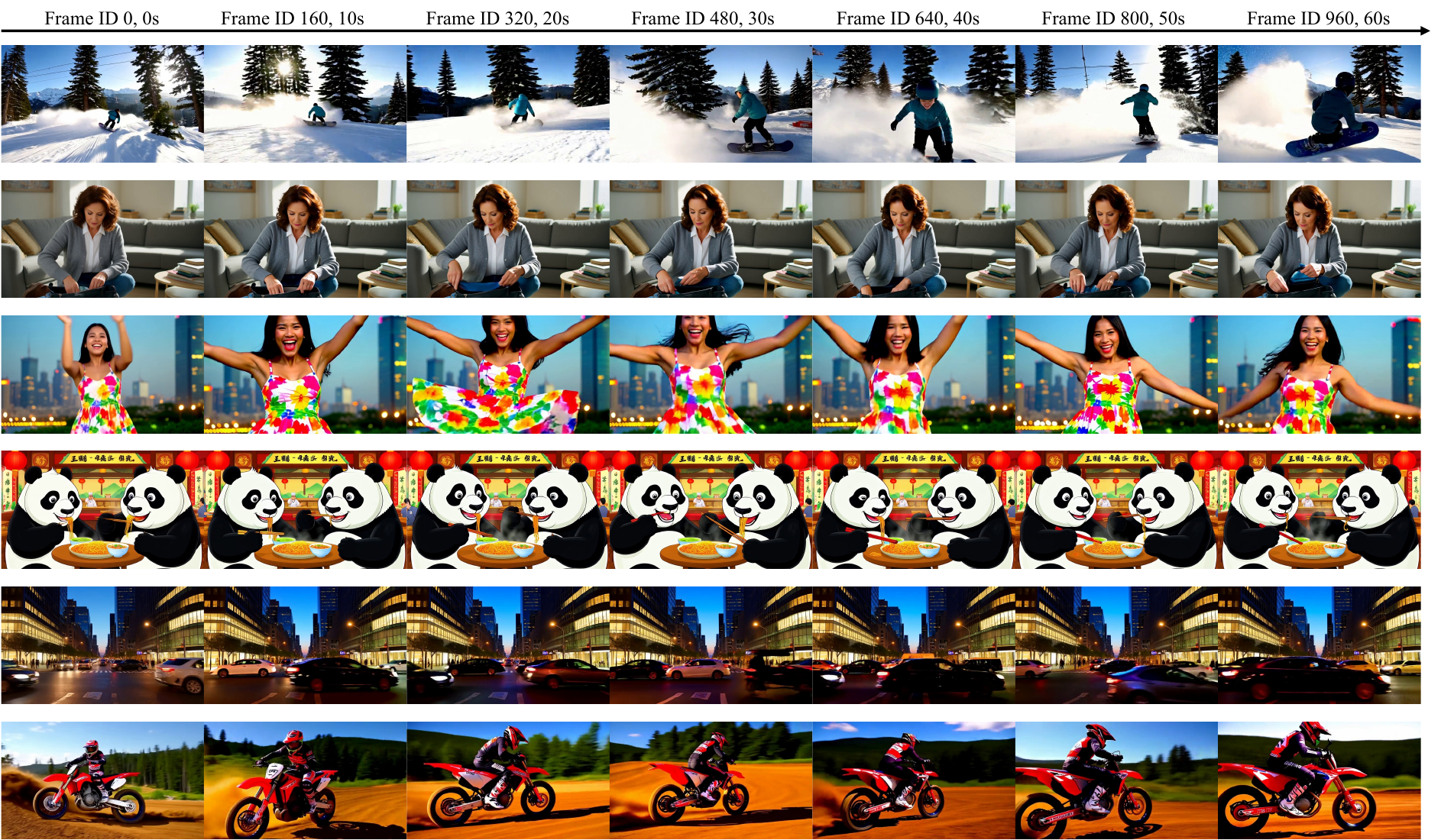}
    \caption{\textbf{Qualitative Results of Context Forcing.} Our method enables minute-level video generation with minimal drifting and high consistency across diverse scenarios.}
    \label{fig:main_demo}
    \vspace{-2mm}
\end{figure*}

\subsection{Context Management System}
\label{sec:long-context-teacher}
Our teacher and student models share an identical architecture; both are autoregressive generative models augmented with a memory module for context retention.
We utilize KV caches to represent the context $X_{1:k}$. To maintain efficiency as the sequence length $k$ grows, we design a KV cache management system inspired by dual-process memory theories. Specifically, the cache $\mathcal{M}$ is partitioned into three functional components: an \emph{Attention Sink}, \emph{Slow Memory} (Context), and \emph{Fast Memory} (Local). Both the student and teacher are equipped with this system.

\noindent \textbf{Cache Partitioning.}
The total cache is defined as the union of disjoint sets:
\[
\mathcal{M} = \mathcal{S} \cup \mathcal{C}_{\text{slow}} \cup \mathcal{L}_{\text{fast}}.
\]
\begin{itemize}
    \item \textit{Attention Sink ($\mathcal{S}$)}: Retains initial $N_s $tokens to stabilize attention, following StreamingLLM~\cite{yang2025longlive, shin2025motionstream}.
    
    \item \textit{Slow Memory ($\mathcal{C}_{\text{slow}}$)}: A long-term buffer of up to $N_c$ tokens, storing high-entropy keyframes and updating only with significant new information.
    
    \item \textit{Fast Memory ($\mathcal{L}_{\text{fast}}$)}: A rolling FIFO queue of size $N_l$, capturing immediate local context with short-term persistence.
\end{itemize}

\noindent \textbf{Surprisal-Based Consolidation.}
Upon generating a new token $x_t$ and enqueuing it into the Fast Memory $\mathcal{L}_{\text{fast}}$, we evaluate its informational value relative to the immediate temporal context. We postulate that tokens exhibiting high similarity to their predecessors carry redundant information (low surprisal), whereas dissimilar tokens indicate significant state transitions or visual changes (high surprisal).

To capture these high-information moments efficiently, we compare the key vector $k_t$ of the current token with that of the immediately preceding token $k_{t-1}$. The consolidation policy $\pi(x_t)$ determines whether $x_t$ is promoted to Slow Memory $\mathcal{C}_{\text{slow}}$:
\begin{equation}
    \pi(x_t) = 
    \begin{cases} 
    \text{Consolidate} & \text{if } \text{sim}(k_t, k_{t-1}) < \tau, \\
    \text{Discard} & \text{otherwise,}
    \end{cases}
\end{equation}
where $\tau$ is a similarity threshold. This criterion ensures that $\mathcal{C}_{\text{slow}}$ prioritizes storing temporal gradients and distinctive events rather than static redundancies. As with standard cache management, if $|\mathcal{C}_{\text{slow}}| > N_c$ after consolidation, the oldest entry is evicted to maintain fixed memory complexity.

\noindent \textbf{Bounded Positional Encoding.}
Unlike standard autoregressive video models~\citep{huang2025self, cui2025self}, where positional indices grow unbounded ($p_t = t \to \infty$), leading to distribution shifts on long sequences, we adopt \emph{Bounded Positional Indexing}. All tokens' temporal RoPE positions are constrained to a fixed range $\Phi = [0, N_s + N_c + N_l - 1]$ regardless of generation step $t$:
\begin{equation}
    \phi(x) = 
    \begin{cases} 
    i \in [0, N_s - 1]  & \text{if } x \in \mathcal{S}, \\
    j \in [N_s, N_c - 1] & \text{if } x \in \mathcal{C}_{\text{slow}}, \\
    k \in [N_c, N_c + N_l - 1] & \text{if } x \in \mathcal{L}_{\text{fast}}.
    \end{cases}
\end{equation}
 This creates a static attention window where recent history (Fast) slides through high indices, while salient history (Slow) is compressed into lower indices, stabilizing attention over long sequences.

\subsection{Robust Context Teacher Training}
\label{sec:teacher_training}
Standard training conditions the model on ground-truth context, but inference relies on self-generated history, creating a distribution shift known as exposure bias. To ensure our Context Teacher provides robust guidance even when the student drifts, we adopt Error-Recycling Fine-Tuning (ERFT)~\citep{li2025stable}. 

Rather than training on clean history $X_{1:k}$, we inject realistic accumulated errors into the teacher's context. We construct a perturbed context $\tilde{X}_{1:k} = X_{1:k} + \mathbb{I} \cdot e_{\text{drift}}$, where $e_{\text{drift}}$ is sampled from a bank of past model residuals and $\mathbb{I}$ is a Bernoulli indicator. The teacher is optimized to recover the correct velocity $v_{\text{target}}$ from $\tilde{X}_{1:k}$. This active correction capability ensures $p_T(\cdot \mid X_{1:k})$ remains a reliable proxy for $p_{\text{data}}$ even when the student's context $X_{1:k}$ degrades.

%% file: sec/4_experiments.tex
\section{Experiments}

\noindent\textbf{Implementation Details.}
We implement the robust context teacher using Wan2.1-T2V-1.3B~\cite{wan2025wan} as the base model. To construct the training dataset, we filter the Sekai~\cite{li2025sekai} and Ultravideo~\cite{xue2025ultravideo} collections to retain high-quality videos exceeding 10 seconds in duration, yielding a total of 40k clips. The robust context teacher is trained for 8k steps with a batch size of 8. During training, frames are sampled uniformly from the 5--20 second interval of the video data to serve as context.

The student model also utilizes the Wan2.1-T2V-1.3B model. In Stage~1, we employ 81-frame video clips from the VidProM~\cite{wang2024vidprom} dataset and train for 600 iterations with a batch size of 64. In Stage~2, which focuses on context distillation, we extend the rollout horizon to video lengths of 10--30 seconds to address short-term memory limitations. This phase is similarly trained on the VidProM dataset for 500 iterations using the same batch size. For both teacher and student models, we set the KV cache size to 21 latent frames, and set $N_s=3, N_c=12,N_l=6, \tau=0.95$. We implement Surprisal-Based Consolidation at 2-chunk intervals. Upon chunk consolidation, we retain only the first latent, effectively extending the context beyond 20s.

\noindent\textbf{Baselines.}
We evaluate our method against three distinct categories of baselines. The first category comprises bidirectional diffusion models, specifically LTX-Video~\cite{hacohen2024ltx} and Wan2.1~\cite{wan2025wan}. The second category includes autoregressive models such as SkyReels-V2~\cite{chen2025skyreels}, MAGI-1~\cite{teng2025magi}, CausVid~\cite{yin2024slow}, NOVA~\cite{deng2024autoregressive}, Pyramid-Flow~\cite{jin2024pyramidal}, and Self Forcing~\cite{huang2025self}. The third category consists of recent methods targeting long video generation within autoregressive frameworks. These include LongLive~\cite{yang2025longlive} with a context length of 3 seconds, Self Forcing++~\cite{cui2025self}, Rolling Forcing~\cite{liu2025rolling} with a context length of 6 seconds, and Infinity-RoPE~\cite{yesiltepe2025infinity} with a context length of 1.5 seconds. Finally we include a long context baseline Framepack~\cite{zhang2025packing} with a context length of 9.2 seconds.

\noindent\textbf{Evaluation.}
We report performance on VBench~\cite{zheng2025vbench} following~\cite{huang2025self, yang2025longlive}. Beyond standard benchmarks, we assess fine-grained consistency using DINOv2~\cite{oquabdinov2}  (structural identity), CLIP-F~\cite{radford2021learning}  (semantic context), and CLIP-T (prompt alignment). To improve robustness against temporal artifacts, we implement window-based sampling: for any timestamp $t$, we compute the average cosine similarity between the first frame ($V_0$) and frames within $[t-0.5s, t+0.5s]$. We average results over five random seeds per prompt to ensure statistical reliability. This approach effectively measures long-term subject and background consistency.

\begin{figure}[tp]
    \centering
    \includegraphics[width=0.95\linewidth]{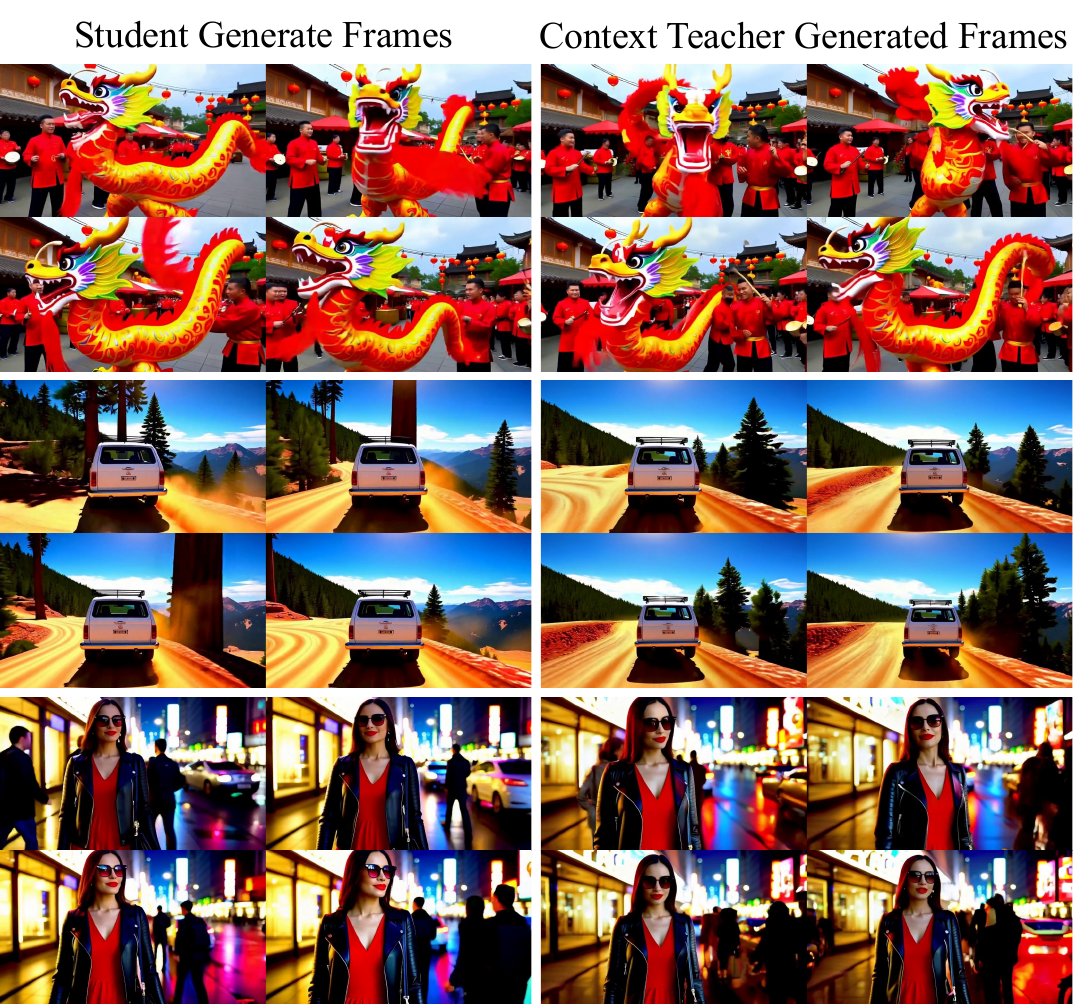}
    \caption{\textbf{Video Continuation with Robust Context Teacher.} Context teacher can generate next segment videos with context generated by student.}
    \label{fig:context_demo}
    \vspace{-4mm}
\end{figure}

\begin{table*}[t]
\centering
\captionof{table}{Single-prompt 60-second long video consistency evaluation.}
\label{tab:long_dino} 
\setlength{\tabcolsep}{6pt}
\renewcommand{\arraystretch}{1.15}
\resizebox{\linewidth}{!}{%
\begin{tabular}{lccccccccccccccccccccc}
\toprule
\textbf{Model} & \textbf{Context Length $\uparrow$} & \multicolumn{6}{c}{\textbf{Dino Score $\uparrow$}} & \multicolumn{6}{c}{\textbf{Clip-F Score $\uparrow$}} & \multicolumn{6}{c}{\textbf{Clip-T Score $\uparrow$}} & \textbf{Background} & \textbf{Subject}   \\
\cmidrule(lr){3-8} \cmidrule(lr){9-14} \cmidrule(lr){15-20}
 & & \textbf{10s} & \textbf{20s} & \textbf{30s} 
 & \textbf{40s} & \textbf{50s} & \textbf{60s} & \textbf{10s} & \textbf{20s} & \textbf{30s} 
 & \textbf{40s} & \textbf{50s} & \textbf{60s} & \textbf{10s} & \textbf{20s} & \textbf{30s} 
 & \textbf{40s} & \textbf{50s} & \textbf{60s}  & \textbf{Consistency$\uparrow$} & \textbf{Consistency$\uparrow$}\\
\midrule
FramePack-F1 & 9.2s & 81.86 & 77.95 & 78.84 & 73.10 & 69.52 & 68.50 & 95.41 & 93.42 & \underline{93.92} & \underline{92.44} & 92.47 & 89.36 & 36.36 & 34.67 & 33.75 & 33.84 & 34.77 & 32.30 & 91.61 & 89.15 \\
LongLive      & 3.0s & \underline{91.25} & \textbf{89.55} & \underline{89.12} & \underline{86.51} & \underline{87.83} & \underline{86.26} & \underline{95.74} & \underline{94.25} & 92.80 & 91.50 & \underline{93.12} & \underline{94.82} & \underline{36.95} & \underline{35.80} & \underline{36.17} & \underline{36.58} & \underline{35.92} & \underline{37.13} & 94.92 & 93.05 \\
Infinate RoPE & 1.5s & 91.18 & 88.17 & 85.37 & 79.80 & 81.10 & 83.72 & 94.09 & 91.71 & 91.13 & 89.71 & 86.11 & 88.88 & 35.26 & 35.03 & 35.88 & 32.56 & 32.29 & 32.28 & 92.42 & 90.11 \\
\rowcolor{yellow!20} \textbf{Ours, teacher} & 20+s & 87.61 & - & - & - & - & - & 95.52 & - & - & - & - & - & 35.93 & - & - & - & - & - & \underline{95.24} & \underline{94.87} \\
\rowcolor{yellow!20} \textbf{Ours, student} & 20+s & \textbf{91.45} & \underline{89.25} & \textbf{89.66} & \textbf{87.45} & \textbf{88.33} & \textbf{87.89} & \textbf{95.82} & \textbf{94.75} & \textbf{95.05} & \textbf{93.88} & \textbf{94.92} & \textbf{95.35} & \textbf{37.12} & \textbf{36.25} & \textbf{36.72} & \textbf{37.15} & \textbf{37.08} & \textbf{37.66} & \textbf{95.95} & \textbf{95.68} \\
\bottomrule
\end{tabular}}
\label{tab:dino-score}
\vspace{-4mm}
\end{table*}

\begin{table*}[tp]
\centering
\caption{Comparison of video generation models across architecture families.}
\resizebox{\linewidth}{!}{%
\begin{tabular}{lcccccccccccc}
\toprule
\textbf{Model} & \textbf{\#Params} & 
\textbf{Throughput (FPS) $\uparrow$} & 
\multicolumn{5}{c}{\textbf{Evaluation scores on 5s $\uparrow$}} & 
\multicolumn{5}{c}{\textbf{Evaluation scores on 60s $\uparrow$}} \\
\cmidrule(lr){4-8} \cmidrule(lr){9-13}
& & & & & & \textbf{Background} & \textbf{Subject} & & & & \textbf{Background} & \textbf{Subject} \\
 & & & \textbf{Total} & \textbf{Quality} & \textbf{Semantic} & \textbf{Consistency} & \textbf{Consistency} & \textbf{Total} & \textbf{Quality} & \textbf{Semantic} & \textbf{Consistency} & \textbf{Consistency} \\

\midrule
\textit{Bidirectional models} \\
LTX-Video & 1.9B & 8.98 & 80.00 & 82.30 & 70.79 & 95.30 & 95.01 &  - & - & - & - & - \\
Wan2.1 & 1.3B & 0.78 & \underline{84.26} & \underline{85.30} & 80.09 & \underline{96.96} & 95.99 &  - & - & - & - & - \\
\midrule
\textit{Autoregressive models} \\
SkyReels-V2  & 1.3B & 0.49 & 82.67 & 84.70 & 74.53 & 96.83 & 96.07 & 70.47 & 75.30 & 51.15 & 89.95 & 84.99 \\
MAGI-1 & 4.5B & 0.19 & 79.18 & 82.04 & 67.74 & 96.83 & 95.83 & 69.87 & 76.12 & 44.87 & 87.76 & 79.46 \\
CausVid  & 1.3B & 17.0 & 81.20 & 84.05 & 69.80 & 95.12 & 95.96 & 71.04 & 76.80 & 48.01 & 89.85 & 86.75 \\
NOVA  & 0.6B & 0.88 & 80.12 & 80.39 & 79.05 & 95.16 & 93.38 & 65.25 & 70.25 & 45.24 & 88.06 & 77.50 \\
Pyramid Flow  & 2B & 6.7 & 81.72 & 84.74 & 69.62 & 96.09 & 96.08 & - & - & - & - & - \\
Self Forcing, chunk-wise & 1.3B & 17.0 & 84.31 & 85.07 & \textbf{81.28} & 95.98 & \underline{96.29} & 71.86 & 77.20 & 50.51 & 87.84 & 83.60 \\
\midrule
\textit{Long autoregressive models} \\
LongLive & 1.3B & 20.7 & \textbf{84.87} & \textbf{86.97} & 76.47 & 96.55 & 95.82 & \textbf{83.64} & \textbf{84.53} & \underline{74.97} & \underline{94.62} & \underline{93.88} \\
Self Forcing++ & 1.3B & 17.0 & 83.11 & 83.79 & \underline{80.37} & - & - & - & - & - & - & - \\
Rolling Forcing & 1.3B & 15.8 & 81.22 & 84.08 & 69.78 & 96.11 & 96.02 & 79.31 & 81.87 & 67.69 & 94.12 & 93.10 \\
Infinity-RoPE & 1.3B & 17.0 & 81.79 & 83.27 & 75.87 & 96.34 & 95.14 & 79.99 & 80.81 & 74.30 & 94.21 & 93.05 \\
\midrule
\rowcolor{yellow!20}  \textbf{Ours, student model} & 1.3B & 17.0 & 83.44 & 84.98 & 77.29 & \textbf{97.38} & \textbf{96.84} & 82.45 & \underline{83.55} & \textbf{76.10} & \textbf{95.34} & \textbf{94.88} \\
\bottomrule
\end{tabular}
\label{tab:model}
}
\end{table*}


\subsection{Video Continuation with Robust Context Teacher}\label{exp:context_continuation} 
To evaluate the context teacher, we feed the teacher model with videos generated by the student model after Stage~1 training. We then assess the consistency of the complete sequence, which comprises the initial context with the generated continuation. Evaluation is performed using 100 text prompts randomly sampled from MovieGenBench~\cite{polyak2024movie}. As illustrated in Figure~\ref{fig:context_demo}, the context teacher effectively synthesizes the subsequent video segment, providing empirical support for Assumptions 1 and 2. Furthermore, we quantitatively evaluate the performance of the context teacher using student-generated videos as input, reporting subject and background consistency on VBench, as well as DINOv2, CLIP-F, and CLIP-T scores. The consistency metrics for the complete 10-second sequence are presented in Table~\ref{tab:long_dino}, further demonstrating that the context teacher consistently produces reliable continuations from student-generated contexts.

\subsection{Text-to-Short Video Generation}


\noindent\textbf{Quantitative Results.}
We quantitatively compare our method against baselines. We evaluate 5-second video generation on the VBench dataset using its official extended prompts. The results summarized in Table~\ref{tab:model} demonstrate that our method achieves performance comparable to the baselines on short video generation.

\subsection{Text-to-Long Video Generation}

\noindent\textbf{Qualitative Results.}
We evaluate our proposed method against baseline models on 60-second video generation, with qualitative results illustrated in Figure~\ref{fig:compare_vis}. By leveraging a slow-fast memory architecture with a KV cache size of 21 and a context span exceeding 20s, our method achieves superior consistency and effectively mitigates content drifting compared to the baselines.

\noindent\textbf{Quantitative Results.}
We evaluate 60-second video generation performance on the VBench with results summarized in Table~\ref{tab:model}, using its offical extened prompts. Additionally, we report DINOv2, CLIP-F, and CLIP-T scores in Table~\ref{tab:dino-score}, using 100 text prompts randomly sampled from MovieGenBench~\cite{polyak2024movie}, following the same experimental protocol as in Section~\ref{exp:context_continuation}. Both tables demonstrate that our method achieves high consistency, particularly during extended video sequences. Notably, while LongLive also achieves competitive scores, qualitative inspection reveals that it frequently exhibits abrupt scene resets and cyclic motion patterns, shown in Figure~\ref{fig:longlive_back} in Appendix.

\subsection{Ablation Studies}

\begin{table}[tp]
\centering
\caption{Ablation study on Slow Memory Sampling Strategy, Context DMD, and Bounded Positional Encoding (evaluated on 60s).}
\label{tab:long} 

\setlength{\tabcolsep}{4pt} 
\renewcommand{\arraystretch}{1.2}
\resizebox{\linewidth}{!}{%
\begin{tabular}{lcccccc}
\toprule
\textbf{Model} & \makecell{Total\\Score $\uparrow$} & \makecell{Quality\\Score $\uparrow$} & \makecell{Semantic\\Score $\uparrow$} & \makecell{Background\\ Consistency $\uparrow$} & \makecell{Subject\\ Consistency $\uparrow$} & \makecell{Dynamic\\ Degree $\uparrow$} \\
\midrule
\multicolumn{7}{l}{\textit{Slow Memory Sampling Strategy}} \\
Uniform sample, interval 1    & 80.82 & 82.20 & \underline{75.32} & 92.45 & 92.10 & 52.15 \\
Uniform sample, interval 2    & \underline{81.11} & \underline{82.61} & 75.12 & 93.12 & 92.85 & \underline{55.30} \\
\midrule
\multicolumn{7}{l}{\textit{Contextual Distillation}} \\
w/o. Contextual Distillation & 80.36 & 82.28 & 72.70 & \underline{93.55} & \underline{93.20} & 48.12 \\
\midrule
\multicolumn{7}{l}{\textit{Bounded Positional Encoding}} \\
w/o. Bounded Positional Encoding        & 73.52 & 75.44 & 65.82 & 84.68 & 79.24 & 27.45 \\
\midrule
\rowcolor{yellow!20} \textbf{Ours} & \textbf{82.45} & \textbf{83.55} & \textbf{76.10} & \textbf{95.34} & \textbf{94.88} & \textbf{58.26} \\
\bottomrule
\end{tabular}}
\vspace{-4mm}
\end{table}

\begin{figure}[tp]
    \centering
    \includegraphics[width=0.95\linewidth]{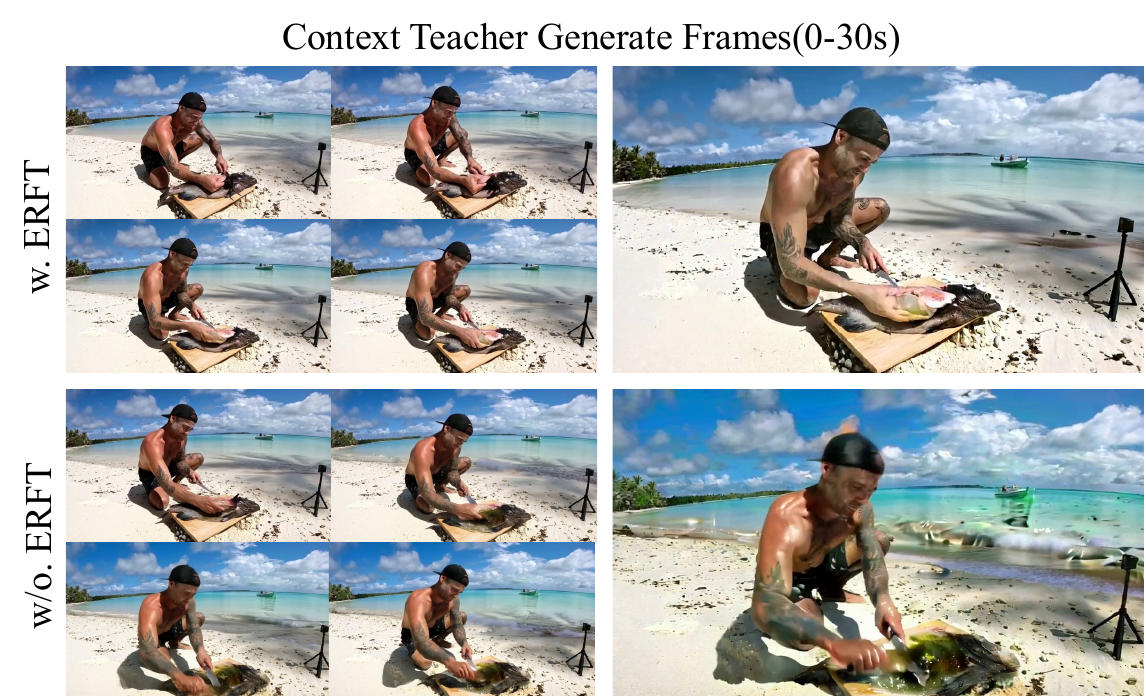}
    \caption{\textbf{Ablation on Error-Recycling Fine-Tuning (ERFT)}. With ERFT, context teacher is more robust to accumulate error.}
    \label{fig:erft}
    \vspace{-4mm}
\end{figure}

\noindent\textbf{Slow Memory Sampling Strategy}
Our method employs a selection strategy based on key-vector similarity to sample context from slow memory. Unlike fixed uniform sampling, this strategy dynamically selects historical chunks that exhibit low similarity to the current generation window, thereby preserving critical semantic information over time. We compare our approach against alternative baselines, specifically uniform sampling with intervals of 1 and 2 chunks. As summarized in Table~\ref{tab:long}, the results demonstrate the effectiveness of similarity-based selection in maintaining long-term consistency.

\noindent\textbf{Context DMD Distillation}
We evaluate the contribution of Contextual Distribution Matching Distillation by comparing our full model against a training-free baseline. In the latter, our context management system is applied directly after Stage~1 training without the DMD process. The results in Table~\ref{tab:long} indicate that removing Context DMD leads to a degradation in both semantic and temporal consistency, highlighting its critical role in enabling coherent, long-horizon video generation.

\noindent\textbf{Error-Recycling Fine-Tuning (ERFT).} We test the context teacher by taking 5s videos from the video dataset as input for autoregressive rollout. As shown in Figure~\ref{fig:erft}, the visualization of 30s generation results indicates that with robust context training, the context teacher produces videos with fewer artifacts. This results in a better distribution for further contextual distillation.

\noindent\textbf{Bounded Positional Encoding.} We investigate the impact of Bounded Positional Encoding by excluding it during inference, with quantitative results presented in Table~\ref{tab:long}. In the absence of this encoding, we observe a significant performance drop in both background stability and subject consistency. This demonstrates its essential role in stabilizing long-range attention and mitigating temporal drift during the generation process.

%% file: sec/5_conclusion.tex
\section{Conclusion}In this work, we introduced \textbf{Context Forcing}, a framework designed to overcome the fundamental \textbf{student-teacher mismatch} in long-horizon causal video generation. By ensuring the teacher model maintains awareness of long-term history, our approach eliminates the supervision gap that limits existing streaming-tuning methods. To handle the computational demands of extreme durations, we proposed a \textbf{Slow-Fast Memory} architecture that effectively reduces visual redundancy. Extensive experiments demonstrate that Context Forcing achieves effective context lengths of 20+ seconds, a $2\text{--}10\times$ improvement over current state-of-the-art baselines. While our method significantly mitigates drifting errors and enhances temporal coherence, the current memory compression strategy still leaves room for optimization regarding information density. Future work can focus on learnable context compression and adaptive memory mechanisms to further improve efficiency and semantic retention for even more complex, open-ended video synthesis.

%% file: sec/6_impact_statement.tex
\section*{Impact Statement}

This paper contributes to the advancement of generative AI by enhancing temporal consistency in long video generation. Our work enables the creation of more coherent and realistic visual sequences, which has significant positive potential in digital storytelling, filmmaking, world model and professional video editing. However, we acknowledge that the ability to generate highly consistent long-form videos also increases the risk of creating sophisticated synthetic media or deepfakes that could be used for misinformation. To mitigate these ethical concerns, we advocate for the integration of digital watermarking and provenance standards in downstream applications. We believe that fostering transparency and developing robust detection mechanisms are essential as video generation technology continues to mature.

%% file: sec/X_suppl.tex
\newpage
\appendix
\onecolumn

\section{Preliminaries}
\noindent \textbf{Causal Autoregressive Models.}  
Causal autoregressive models generate videos at the frame or short-chunk level ($X_t$) while enforcing strict temporal causality. Methods such as CausVid~\cite{yin2024slow} and Self-Forcing~\cite{huang2025self} adopt block-wise causal attention, allowing bidirectional self-attention within each chunk $X_t$ but restricting information flow across chunks. Video generation is formulated as $P(X_t \mid X_{<t})$. In Self-Forcing, the student model is stochastically conditioned on its own generated outputs $\hat{X}_{<t}$ during training. These models typically employ Distribution Matching Distillation (DMD)~\cite{yin2024one} to distill knowledge from a bidirectional teacher into a causal student.

\section{Visual artifacts in LongLive.}\label{supp:longlive}
\begin{figure}[tp]
    \centering
    \includegraphics[width=0.9\linewidth]{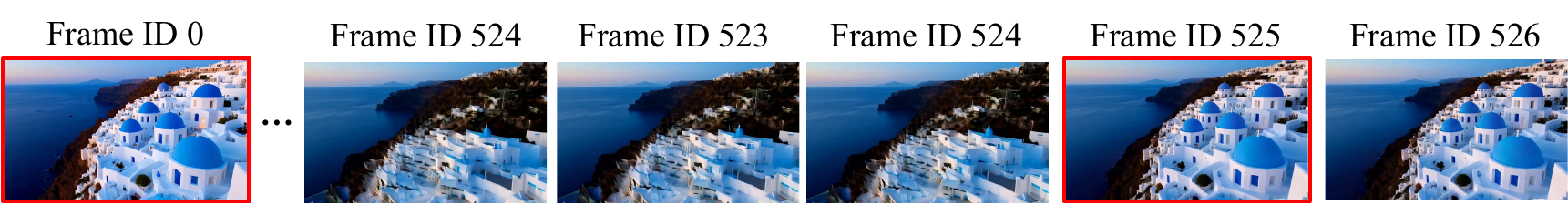}
    \caption{\textbf{Visual artifacts in LongLive.} The model exhibits a sudden flashback artifact, where the video abruptly resets to the initial frame after 524 frames, disrupting temporal continuity.}
    \label{fig:longlive_back}
\end{figure}

While LongLive achieves respectable quantitative scores, we observe that it frequently suffers from abrupt scene resets and repetitive, cyclic motion patterns, as illustrated in Figure~\ref{fig:longlive_back}.

\section{Algorithm of Context Forcing.}

Algorithm block of context forcing.

\begin{algorithm}[tp]
\caption{Contextual DMD}
\label{alg:contextualdmd}
\begin{algorithmic}[1]
\Require Denoise timesteps $\{t_1,..,t_T\}$
\Require Pre-trained teacher $s_{real}$
\Require Checkpoints from stage 1, student score function $s_{fake}$, AR diffusion model $G_{\phi}$
\Require Text prompt dataset $\mathcal{D}$, rollout decay step $s_d$, rollout range $(L_0, L_1)$, context window $c$, teacher length $l$, local attention size $a$
\State \textbf{Initialize}, step $ s = 0$
\State \textbf{Initialize} model output $X \leftarrow []$
\State \textbf{Initialize} KV cache $C\leftarrow[]$
\While{training}
    
    \State \textbf{Sample} prompt $p\sim \mathcal{D}$

    \State \textbf{Sample} rollout length $L = \text{Uniform}(L_0 , \frac{s}{s_d}\times (L_1-L_0)+L_0+1)$

    \State \textbf{Sample} random exit $r = \text{Uniform}(1, 2, ... ,T)$

    \For{$i=1,...,L$}

        \State \textbf{Initialize} $x_t^i\sim\mathcal{N}(0,\mathrm{I})$

        \If{$L-r-l\leq i<L-l $}
        
        \State $r^\prime = T $
        \Else
        
        \State $r^\prime = r $
        \EndIf

        \For{$j=1,...,r^\prime$}

        \If{$j = r^\prime$}
        \State Enable gradient computation
        \State $ \hat{x}_0^i\leftarrow G_\phi(x_{t_j}^i,t_j,C)$ 
        \State $X.\texttt{append}(\hat{x}_0^i)$
        \State Disable gradient computation
        \State $C\leftarrow G^{C}_\phi (\hat{x}_0^i, 0, C)$

        \Else
        \State Disable gradient computation

        \State $ \hat{x}_0^i=G_\phi(x_{t_j}^i,t_j,C)$ 
        \State \textbf{Sample} $\epsilon \sim \mathcal{N}(0, \mathrm{I})$
        \State Set $x^i_{t_{j-1}} \leftarrow \text{addnoise}(\hat{x}^i_0, \epsilon, t_{j-1})$

        \EndIf

        \EndFor
        \State context video $v_c=X[L-r-l:L-l]$, target noise $v_t=\texttt{addnoise}(X[L-l:],t)$
        \State Compute Contextual DMD Loss with  $s_{fake}(v_t,t,v_c)$ and $s_{real}(v_t,t,v_c)$
    \EndFor  

\EndWhile
\end{algorithmic}
\end{algorithm}